\RequirePackage{fix-cm}
\documentclass[twocolumn]{svjour3}          
\smartqed  
\usepackage{graphicx}
\usepackage{mathptmx}      
\graphicspath{{"./"},{"Figures/"}}
\usepackage{xcolor}
\usepackage[pdflang={en-US},pdftex,breaklinks=true]{hyperref}
\usepackage{color}
\hypersetup{
    colorlinks=true,
    breaklinks=true,
    linkcolor=blue,
    filecolor=magenta,      
    urlcolor=cyan,
    citecolor=darkgray,
    pdftitle={Deep Learning of Individual Aesthetics},
    pdfpagemode=FullScreen,
}

\journalname{Neural Computing and Applications}
\begin{document}

\title{Deep Learning of Individual Aesthetics 
}


\author{Jon McCormack       \and
        Andy Lomas 
}


\institute{Jon McCormack (corresponding author) \at
              SensiLab, Monash University, Australia \\
              \email{Jon.McCormack@monash.edu}           
           \and
           Andy Lomas \at
              Goldsmiths, University of London, UK \\
              \email{a.lomas@gold.ac.uk}
}

\date{Received: 15 June 2020 / Accepted: 22 September 2020}

\maketitle

\begin{abstract}
Accurate evaluation of human aesthetic preferences represents a major challenge for creative evolutionary and generative systems research. Prior work has tended to focus on feature measures of the artefact, such as symmetry, complexity and coherence. However, research models from Psychology suggest that human aesthetic experiences encapsulate factors beyond the artefact, making accurate computational models very difficult to design. The interactive genetic algorithm (IGA) circumvents the problem through human-in-the-loop, subjective evaluation of aesthetics, but is limited due to user fatigue and small population sizes. In this paper we look at how recent advances in deep learning can assist in automating personal aesthetic judgement. Using a leading artist's computer art dataset, we investigate the relationship between image measures, such as complexity, and human aesthetic evaluation. We use dimension reduction methods to visualise both genotype and phenotype space in order to support the exploration of new territory in a generative system. Convolutional Neural Networks trained on the artist's prior aesthetic evaluations are used to suggest new possibilities similar or between known high quality genotype-phenotype mappings. We integrate this classification and discovery system into a software tool for evolving complex generative art and design.

\end{abstract}

\section{Introduction}
\label{intro}

Computational evolutionary methods can support human art\-ists and designers in exploring the aesthetic possibilities of complex generative systems \cite{Bentley1999,BentleyCorne2002,McCormack2005a}. However, the majority of evolutionary algorithms used for art and design rely on being able to compute a fitness measure of phenotype aesthetics. Devising formal aesthetic measures is a long-standing, but generally illusive quest in evolutionary computing and psychology research \cite{Birkhoff1933,Humphrey1973,Johnson2019}.

As a way of circumventing the formalisation of an aesthetic measure, the Interactive Genetic Algorithm (IGA) has long been used by artists and researchers since first being devised by Dawkins in the mid 1980s \cite{Dawkins1986,Sims1991,Todd1991,Todd1992,McCormack1992,Rowbottom1999,McCormack2019cc}. A key advantage of the IGA is that it puts a ``human in the [evolutionary] loop'', substituting formalised fitness evaluation for human judgement. To evolve a visual form, the user simultaneously assesses or compares a small population (typically around 16-25 individuals) from a single parent (if offspring are generated by mutation only) or parents (if crossover is also used) and either ranks or selects the most aesthetically interesting individuals for the breeding pool of parents in the next generation. The process repeats until a satisfactory form is found, or the user runs out of patience.

The algorithm arose to circumvent the difficulty in developing generalised fitness measures for ``subjective'' criteria, such as personal aesthetics or taste. Hence the IGA found favour from many artists and designers, keen to exploit the powerful search and discovery capabilities offered by evolutionary algorithms, but unable to formalise their aesthetic judgement in computable form. Nonetheless, the limitations of the IGA are well known: the user quick\-ly tires or fatigues, limiting the number of generations for evolution; only a small number of offspring can be effectively compared in each generation, keeping the population size very low; users do not necessarily have a strong understanding of the underlying design space, making aesthetic evaluation inconsistent and exploration of the design space limited and ad-hoc \cite{Takagi2001}.

Over the years, the research community has proposed many new theories and measures of aesthetics, with research from both the computational aesthetics (CA) and psychology communities \cite{Johnson2019}. Despite much effort and many advances, a computable, universal aesthetic measure remains an open problem in evolutionary music and art research \cite{McCormack2005a}. 

One of the reasons for this is the psychological nature of aesthetic judgement and experience. In psychology, a detailed model of aesthetic appreciation and judgement has been developed by Leder and colleagues \cite{Leder2004,Leder2014}. This model describes the interactions between various components that integrate into an aesthetic experience and lead to an aesthetic judgement and aesthetic emotion. The model includes perceptual aesthetic properties, such as symmetry, complexity, contrast, and grouping, but also social, cognitive, contextual and emotional components -- all of which contribute significantly to forming an overall aesthetic judgement. A key element of Leder's revised model \cite{Leder2014} is that it recognises the influence of a person's affective state on many components and that aesthetic judgement and aesthetic emotion co-direct each other.

One of the consequences of this model is that any full computational aesthetic measure must take into account the interaction between cognition and affect in the viewer, in addition to other factors such as prior knowledge and experience, the viewing context and deliberate (as opposed to automatic) formulations regarding cognitive mastering, evaluation and social discourse. In sum, factors that are extremely difficult or impossible for current computational models to adequately accommodate.

How then can we progress human-computer collaboration that involves making aesthetic judgements if fully developing a machine-implementable model remains illusive? One possible answer lies in teaching the machine both tacit and learnt knowledge about an individual's personal aesthetic preferences so that the machine can assist a person in creative discovery. The machine provides assistance only, it does not assume total responsibility for aesthetic evaluation or artefact production. Instead it can be used for filtering or suggesting based on learnt measures of individual aesthetics or features.

In this paper we investigate the use of several machine learning (ML) methods, including dimension reduction algorithms and the use of convolutional neural networks (CNNs) as custom classifiers, to assist digital artists in navigating and searching the large design spaces of modern evolutionary generative art systems. The overall aim is for the computer to learn about an individual artist's aesthetic preferences and to use that knowledge to assist them in finding more appropriate phenotypes. ``Appropriate'' in the sense that they fit the artist's conception of high aesthetic value, or that they are in some category that is significant to the artist's creative exploration and partitioning of a design space. For the experiments described in this paper, we worked with real artistic data provided by the second author to give our study ecological validity \cite{Brunswik1956}. The data is unique in the sense that it contains the generative parameters (genotype) and the final creative results (phenotype) along with artist-assigned aesthetic rankings and visual categorisations for 1,774 individuals.

The remainder of this paper is structured as follows: after looking at related work in Section \ref{s:relatedWork}, Section \ref{s:exploringSpace} looks at the design space and details of the artistic datset used. Section \ref{ss:imageAnalysis} examines the relationship between measurable aspects of images -- such as entropy, complexity and fractal dimension -- and personal aesthetics. Prior studies have suggested these measures play a significant role in aspects of visual aesthetics (e.g.~\cite{forsythe2011predicting}). Our analysis shows that while some measures, such as complexity, have reasonably good correlation to the artist's personal aesthetic measure, they are insufficient alone to completely replace it, missing critical visual aspects that differentiate visual forms in the image dataset tested.

Next we explore the use of dimension reduction methods to visualise both genotype and phenotype space (Sections \ref{ss:genotypeVis} and \ref{ss:phenotypeVis}). These maps assist us in understanding the structure of the design space we are exploring. The results show that the genotype space is largely unstructured, whereas the phenotype space does have clearly discernible structures that correspond to the artist's aesthetic preferences. To visualise phenotype space we use a standard image classifier (ResNet-50) without additional training. The results confirm that the network is able to distinguish visually important characteristics in the dataset, leading us to further explore the use of deep learning image classifiers.

Thus in Section \ref{s:learningPreferences} we re-train ResNet-50 CNN classifiers on the artist's images, to predict aesthetic ratings and categorisations from new images. The resultant networks are able to predict the artist's aesthetic preferences with high accuracy (e.g.~87.0\% in the case of categorisation). Given this success, we next train a neural network on the genotype data to see if it is able to predict similarly to the phenotype data. Section \ref{ss:genotypeSpace} describes the experiments using a Tabular model network built from scratch. While not as successful as the ResNet-50 based phenotype networks (achieving an accuracy of 68.3\%), the network is significantly better than previous methods used in the artist's system, such as k-nearest neighbour prediction.

In Section \ref{ss:discussion} we show how these predictors can be integrated into evolutionary design software, providing a visual interface with levels of prediction in genotype space for generative evolutionary systems with high numbers of generative parameters.

Lastly, in Section \ref{s:conclusion} we summarise the results and briefly discuss future work in this area, looking at ways in which mapping and learning about both genotype and phenotype space can inspire a search for new phenotypes that share, blend or interpolate visual features from two or more different categories of known examples. These ap\-proach\-es aim to eliminate the user fatigue and other limitations common with traditional IGA approaches. 

\section{Related Work}
\label{s:relatedWork}
In recent years, a variety of deep learning methods have been integrated into evolutionary art and design systems. Blair \cite{Blair2019} used adversarial co-evolution, evolving images using a GP-like system alongside a LeNet-style Neural Network critic. Blair's work sought to generate images that matched known objects, whereas our system is designed to assist in exploration of a design space without knowing specific targets in advance.

Bontrager and colleagues \cite{Botranger2018} describe an evolutionary system that uses a Generative Adversarial Network (GAN), putting the latent input vector to a trained GAN under evolutionary control, allowing the evolution of high quality 2D images in a target domain.

Elgammal and colleagues used a variant of GANs to create an ``art generating agent'' that attempts to synthesise art with a constrained level of novelty from what it has previously produced \cite{Elgammal:2017aa}. The agent also tries to increase ``stylistic ambiguity'' to avoid stylistic repetition. Like many GAN-Art experiments, the model was trained on sample images from the Western canon of classical and modernist painting, so in the sense of O'hear \cite{Ohear1995} the agent conveys only parasitic meaning as ``art''. In contrast, the neural networks described here are trained on a single artist's work and our goal is not to synthesise derivative works from that sample, but to use the network's vision system as a visual discriminator of personal aesthetic preference. 

More closely related is the work of Singh et al. \cite{Singh2019}, who used the feature vector classifier from a Convolutional Neural Network (CNN) to perform rapid visual similarity search. Their application was for design inspiration by rapid\-ly searching for images with visually similar features from a target image, acquired via a smartphone camera. The basis of our method is similar in the use of using a large, pre-trained network classifier (such as ResNet-50) to find visual similarity between generated phenotype images and a database of examples, however our classifier is re-trained on artist-specific datasets, increasing its accuracy in automating personal aesthetic judgement.

Work by Colton et al. \cite{Colton2020} experimented with a number of image classification networks (ResNet, MobileNet and SqueezeNet) to drive an evolutionary art system for causal creation towards categories that were classified with high confidence by the network. As these classification networks were trained on the ImageNet database of real images \cite{deng2009imagenet}, the abstract images generated by this system required active and considered viewing to find the connection between their visual appearance and the neural network's classification. This method was able to find images that evoked interest due to the visual puzzle of trying to find why a generative image ``looked like'' a real object to the network.  Our system differs in that we retrain the network to classify according to individual aesthetic preferences and categories rather than exploiting the gap between a generative system and its classification according to a database of ``real'' images.

\section{Exploring Space in Generative Design}
\label{s:exploringSpace}

\begin{figure*}
\includegraphics[width=\textwidth]{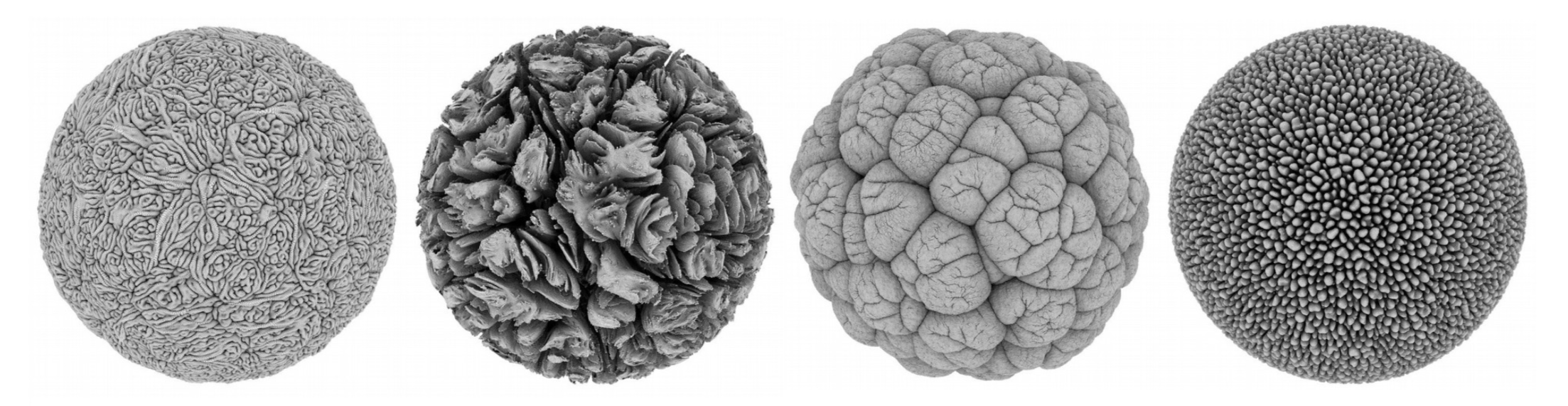}
\caption{Example cellular forms generated by Lomas' cellular morphogenesis algorithm} \label{f:exampleForms}
\end{figure*}

For the experiments described in this paper, we worked with a dataset of evolutionary art created by computational artist Andy Lomas. Lomas works with developmental morphogenetic models that grow and develop via a bespoke cellular development process. Details of the technical mechanisms of his system can be found in \cite{Lomas2014}. A vector of 12 real valued parameters determines the conditions and properties of simulated cell growth and development. The simulation begins from a single cell that repeatedly splits into new cells, growing over time into a complex 3D form that can often involve more than one million cells. The range of forms is quite varied, Figure \ref{f:exampleForms} shows a small selection of samples. In exploring the idea of machine learning of personal aesthetics, we wanted to work with a real, successful artistic system\footnote{Lomas is an award winning computer artist who exhibits internationally, see his website  \url{http://www.andylomas.com}}, rather than an invented one, as this allows us to understand the ecological validity \cite{Brunswik1956} of any system or technique developed. Ecological validity requires the assessment of creative systems in the typical environments and contexts under which they are actually experienced, as opposed to a laboratory or artificially constructed setting. It is considered an important methodology for validating research in the creative and performing arts \cite{Jausovec2011}.

\subsection{Generative Art Dataset}
\label{ss:dataset}

\begin{figure}[htbp]
\begin{center}
\includegraphics[width=0.48\textwidth]{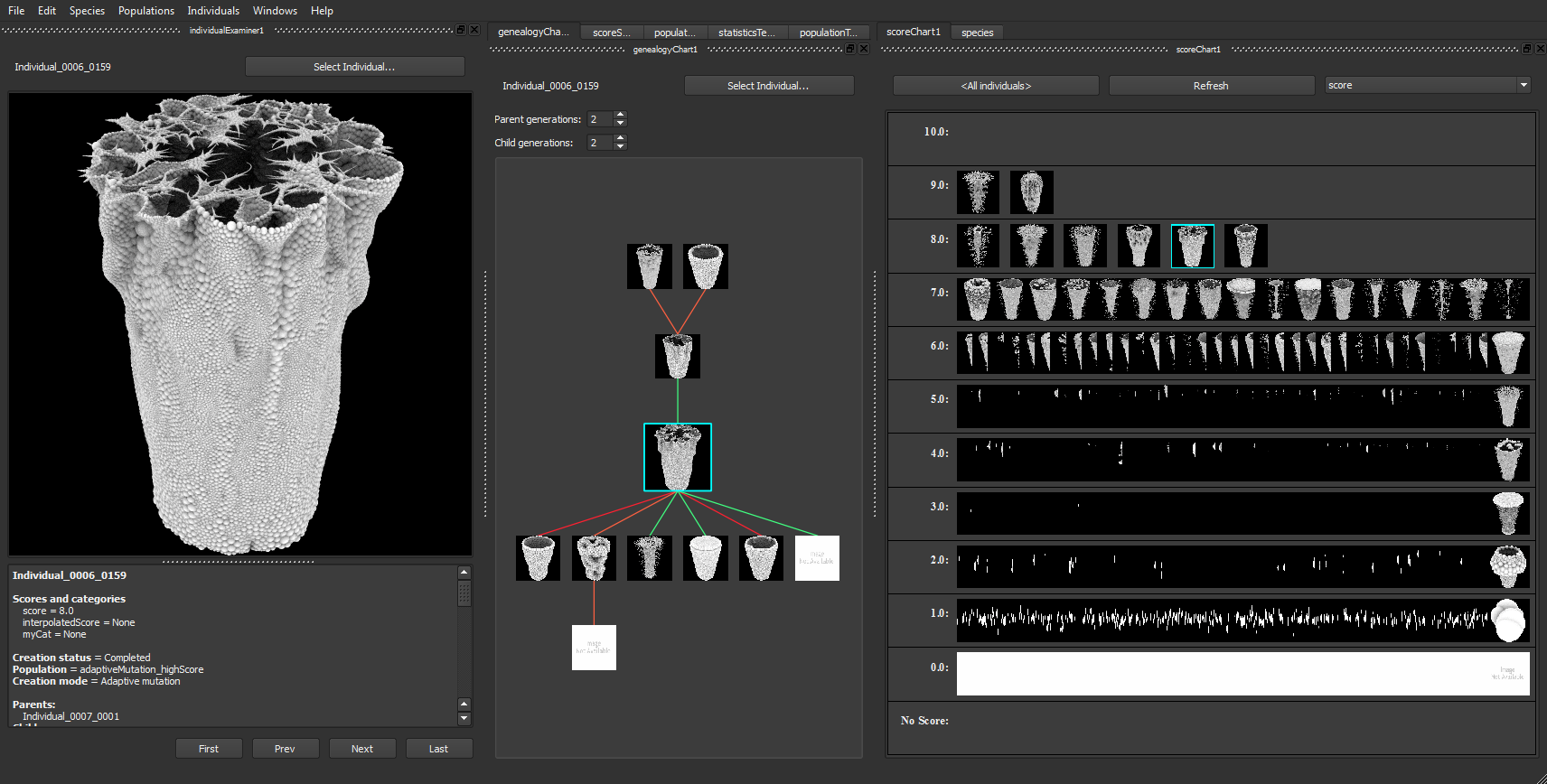}
\end{center}
\caption{The main user interface of \emph{Species Explorer}, showing currently evolved phenotype (left), a genealogy chart for the current individual (middle) and ranking and classification of phenotypes (right).} \label{f:SpeciesExplorerUI}
\end{figure}

The Lomas dataset used consisted of 1,774 images, each generated by a developmental form generation system dri\-ven by the software \emph{Species Explorer} \cite{Lomas2016,Lomas2018}. This software (Figure~%
\ref{f:SpeciesExplorerUI}) is the front-end to the generative developmental system, used for assigning ratings, categories and driving interactive evolution. The system allows a variety of methods to be used to select individuals to be created in the next generation, including simple random sampling of genotype parameters, evolutionary methods such as mutation and cross-breeding of parents picked using fitness functions (IGA), and Monte Carlo selection using simple nearest neighbour based prediction of fitness at new candidate points in the genotype space. It is also worth mentioning that \emph{Species Explorer} was designed as a general graphical front end for generative art systems and the software can support any manor of generative system, provided it uses a genotype to phenotype mapping process.

Each image generated using \emph{Species Explorer} is a two-dimensional rendering of a three-dimensional form that has been algorithmically grown based on 12 numeric parameters (the ``genotype''). The genotype is fed into the form generation system, which simulates the growth, development and division of cells into a 3D form. After a fixed number of time steps the growth is halted and the 3D form is then rendered by the system to create a 2D perspective image (the ``phenotype'').
Rendering involves complex lighting and shading calculations, which impact the visual aesthetics of the image, hence this is the output considered in \emph{Species Explorer} and used by the artist to determine rating scores and categories, described below. As the 2D images, not the raw 3D models are evaluated by the artist, we perform our analysis similarly, using the 2D image renderings as a basis for studying the design space of the system. 

The dataset contains a numeric aesthetic rating score for each form (ranging from 0 to 10, with 1 the lowest and 10 the highest, 0 meaning a failure case where the generative system terminated without generating a form or the result was not rated). These ratings were all performed by Lomas, so represent his personal aesthetic preferences. Rating visual form in this manner is an integral part of using the Species Explorer software, with the values in the dataset created over several weeks as he iteratively generated small populations of forms, rated them, then used those rankings to influence the generation of the next set of forms.

Lomas also developed a series of stylistic categorisations that loosely describe the visual class that each form fits into. This categorisation becomes useful for finding forms \emph{between} or \emph{outside} current categories, discussed in Section \ref{ss:genotypeSpace}. Category labels included ``brain'' (317 images), ``mess'' (539 images), ``balloon'' (169 images), ``animal'' (104 images), ``worms'' (53 images) and ``no growth'' (154 images). As each form develops through a simulated growth process, some of the genomes fail to generate much at all, leading to images that are ``empty'' (all black) or with just a small number of spherical cells. There were 251 empty images, leaving 1,523 images of actual forms. Even though the empty images are not visually interesting, they still hold interesting data as their genomes result in non-viable forms. Most of the category data (1,421 images) had been created at the same time as when Lomas was working on the original Cellular Forms series. The remaining 353 images were categorised by Lomas as part of this research.\footnote{We note that Lomas' classification is purely visual but remarkably consistent. When shown a ``test set'' of 100 of the originally classified images without classification labels, he was able to reclassify them with 97\% accuracy to his original classification.} The full dataset used in all the experiments described in this paper is available online at: \url{https://github.com/SensiLab/Andy-Lomas-Generative-Art-Dataset.git}.

\subsection{Understanding the Design Space}
\label{ss:designSpace}

As a first step in understanding the design space we analysed relationships between genotype, phenotype and personal aesthetic ranking and categorisation from the artist. We first applied image analysis methods to the 2D images, looking for correlations between aesthetic rating or categorisation and overall image properties. The aim of this analysis was to better understand if any common factors, such as symmetry, complexity or morphological structure were influential on personal aesthetic decisions, as these factors are commonly used in general aesthetic measures \cite{Johnson2019}. The results of this analysis is presented in Section \ref{ss:imageAnalysis}.

We then used a variety of dimension reduction algorithms to visualise the distribution of both genotype and phenotype space to see if there was any visible clustering related to either aesthetic ranking scores or categories. We experimented with a number of different algorithms, including t-SNE \cite{maaten2008visualizing}, UMAP \cite{mcinnes2018umap} and Variational Autoencoders \cite{makhzani2015adversarial}, to see if such dimension reduction visualisation techniques could help artists better understand relationships between genotype and categories or highly ranked species. The results of this analysis are presented in Sections \ref{ss:genotypeVis} and \ref{ss:phenotypeVis}.

\subsection{Image Analysis}
\label{ss:imageAnalysis}
We applied a number of image analysis techniques to the entire dataset and looked for correlations between these image measurements and the artist assigned aesthetic score or category. Prior research has proposed image complexity measures, for example, as reasonable proxies for visual beauty \cite{forsythe2011predicting}. We looked at the relationship between several image measures and the aesthetic scores assigned to each image by Lomas. The results are summarised in Table \ref{tab:imageCorrelations}.

\begin{table*}
\caption{Pearson's correlation coefficient values between image measurements and Lomas' aesthetic score. Algorithmic complexity (bold) has the highest correlation with aesthetic score. In all cases $p$-values are $< 1 \times 10^{-10}$}.
\label{tab:imageCorrelations}
\begin{center}
\begin{tabular}{r|cccccccc}
\hline\noalign{\smallskip}
 & Entropy & Energy & Contours & Euler & AComplex & SComplex & FDim & Score  \\
\noalign{\smallskip}\hline\noalign{\smallskip}
Entropy & 1 &  &  & & & & & \\ 
Energy & -0.99 & 1 & & & \\
Contours & 0.43 & -0.38 & 1 & & \\
Euler & -0.42 &  0.37 & -0.99 &  1 & \\
AComplex & 0.97 & -0.95 & 0.50 & -0.50 &  1 \\
SComplex & 0.92 & -0.87 & 0.66 & -0.66 & 0.94 & 1 \\
FDim & -0.35 & 0.45 & 0.29 & -0.30 & -0.16 & -0.05 & 1 \\
Score & 0.63 &  -0.59 & 0.54 & -0.54 & \textbf{0.76} & 0.68 & 0.28 & 1 \\
\noalign{\smallskip}\hline
\end{tabular}
\end{center}
\end{table*}

\begin{figure}[htbp]
\begin{center}
\includegraphics[width=0.48\textwidth]{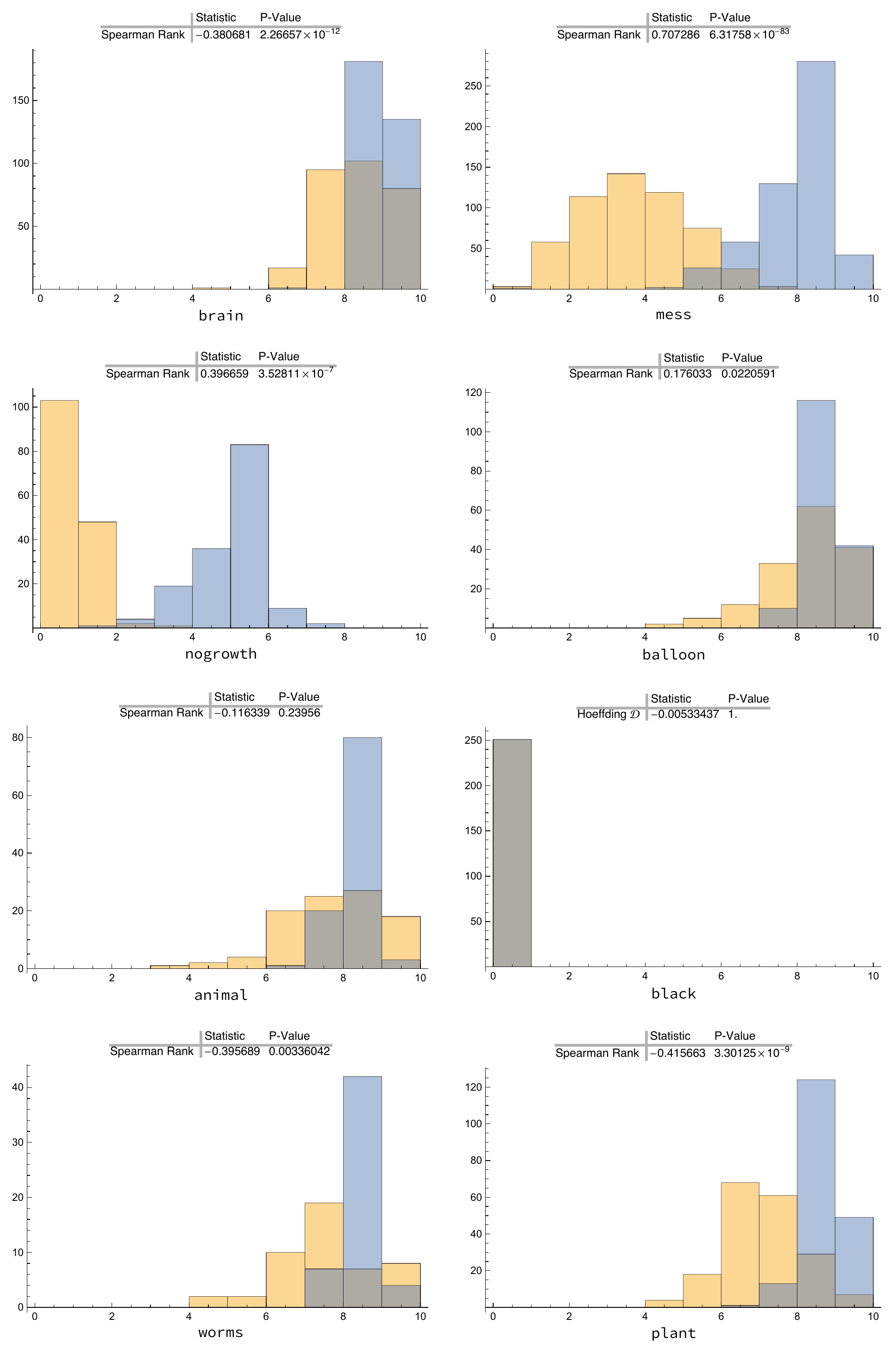}
\end{center}
\caption{Histogram distributions of the aesthetic ranking score (yellow) and image algorithmic complexity (blue) for each form category. Independence testing using Spearman's $\rho$ is shown for each category, with the exception of ``black'' which has scores of all 0} \label{f:ScoreEntropyHisto}
\end{figure}

\emph{Entropy} represents the image data entropy, which can be considered a type of complexity measure. As can be observed in the table, there is moderate positive correlation for this measure.
Similarly \emph{Energy} is the data energy of the image, \emph{Contours} is the number of lines required to describe component boundaries  detected in the image, and \emph{Euler} is the morphological Euler number of the image (effectively a count of the number of connected regions minus the number of holes). The morphological measures (\emph{Contours}, \emph{Euler}) were used as many of the high rated images have detailed regions of complex contours.

Recently some new measures of image complexity have appeared in the literature \cite{Lakhal2020}. Following this method we computed the algorithmic complexity  (\emph{AComplex}) and structural complexity (\emph{SComplex}) of each image using the techniques described in \cite{Lakhal2020}. These measures try to capture the visual complexity in an image using information complexity (compression ratio) in the case of \emph{AComplex} and feature complexity with a form of high-pass filtering (\emph{SComplex}). Structural complexity was computed with a scale radius, $r_{cg} = 5$ and threshold $\delta = 0.23$. As shown in Table \ref{tab:imageCorrelations}, algorithmic complexity had the highest correlation to score. 

We also measured the fractal dimension (\emph{FDim}) of each image using the box-counting method \cite{forsythe2011predicting}. While past analysis of art images has demonstrated relationships between fractal dimension and aesthetics, interestingly in this analysis fractal dimension did not seem a good indicator of Lomas' aesthetic ranking. This is likely because the images do not exhibit traits common to fractals such as self-similarity at multiple scales.

We examined the relationship between these measures and the artist assigned categories. Figure \ref{f:ScoreEntropyHisto} shows histograms of the score (yellow) and image algorithmic complexity measure (blue) for each category along with independence tests for each distribution using Spearman's $\rho$. As the figure shows, the AComplex measure performs better in some categories, such as ``balloon'' and ``animal'', and overall is a reasonable predictor of category and score. Note that the ``black'' class is a special case, with all scores 0, so here for informational purposes we calculate Hoeffding's $D$.

\begin{figure*}[tb]
\begin{center}
\includegraphics[width=0.48\textwidth]{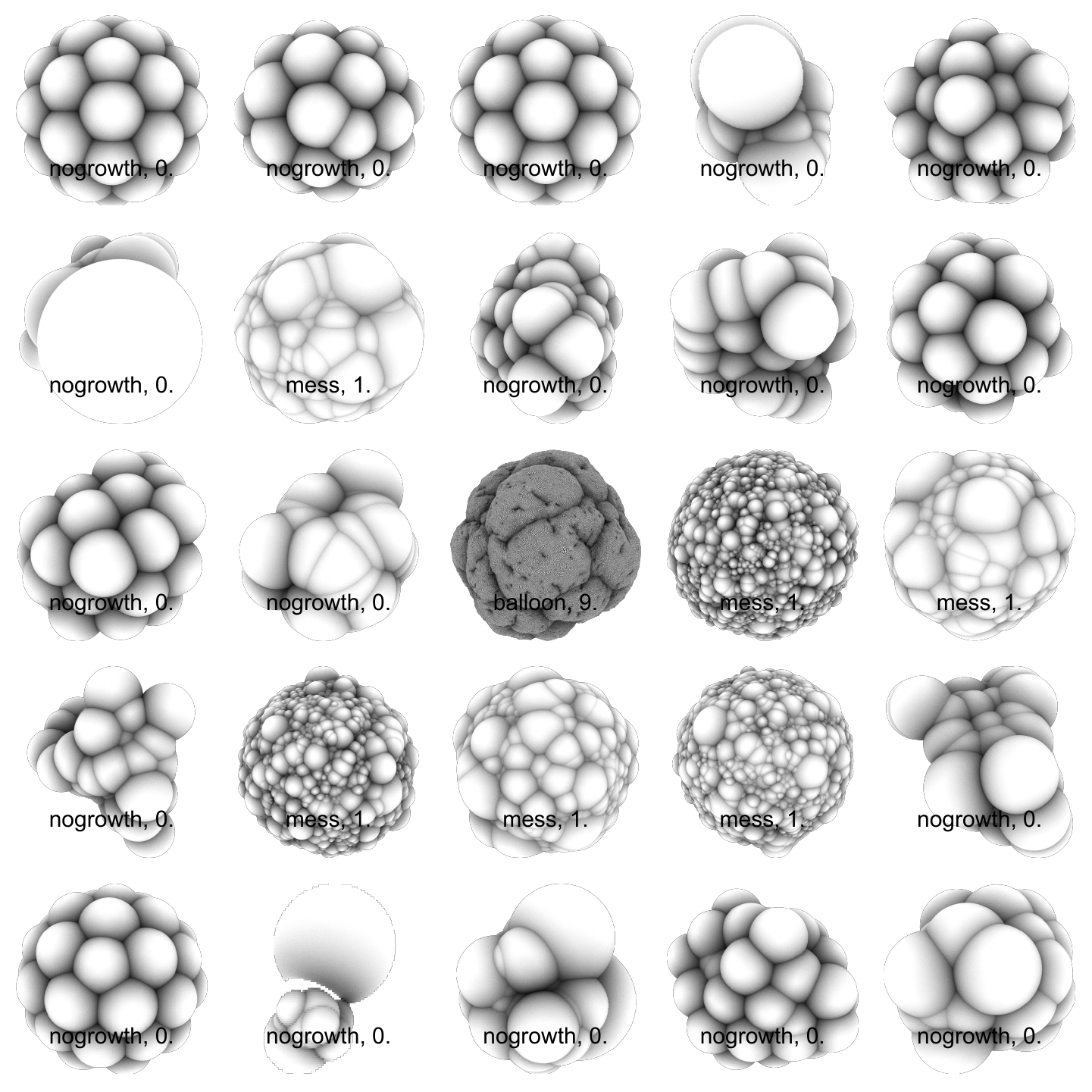}
\includegraphics[width=0.48\textwidth]{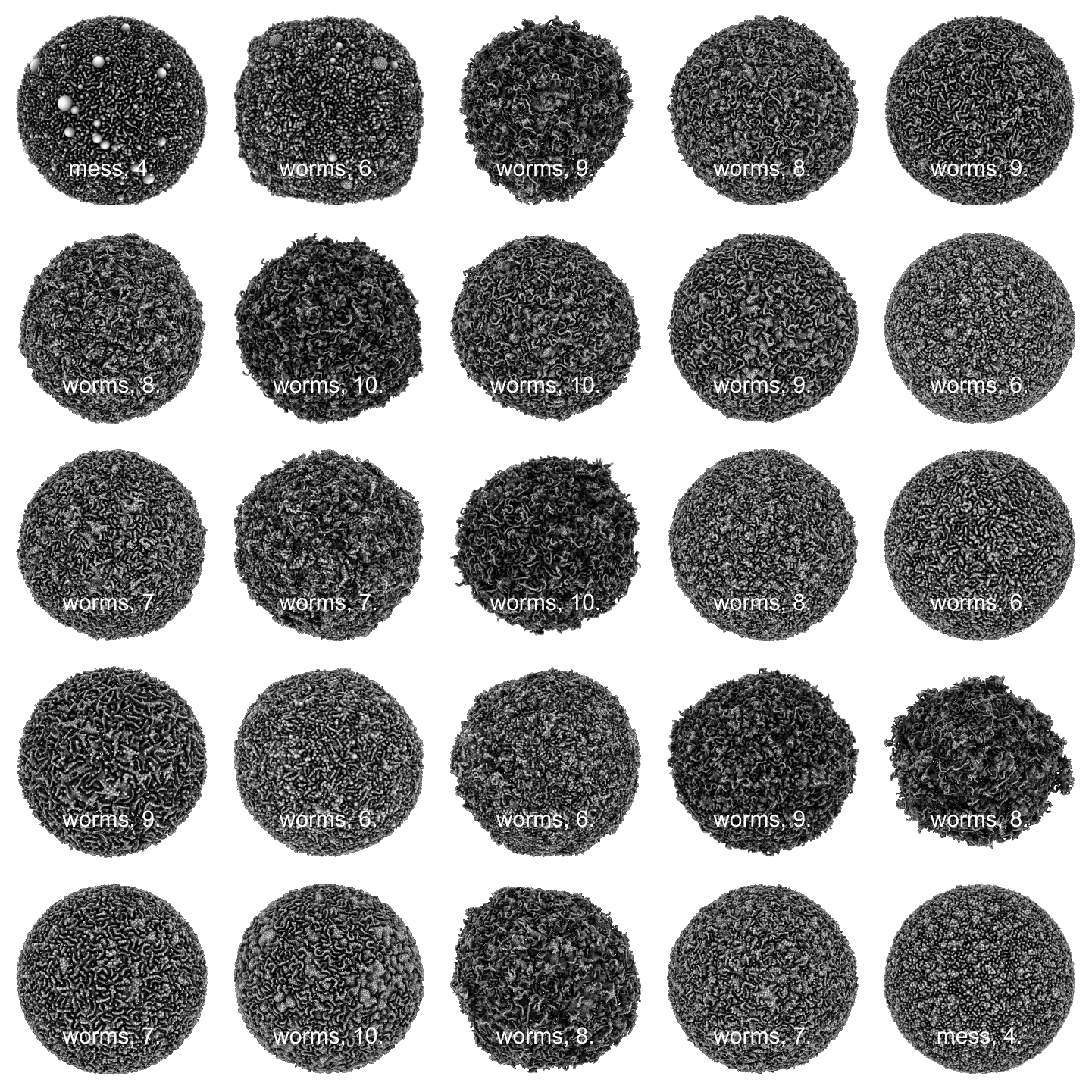}
\end{center}
\caption{Images of the 25 lowest (left) and highest (right) Euler measurements in the dataset. The artist assigned category and score are shown with each image.} \label{f:MENScores}
\end{figure*}

While such measures give a reasonable visual ordering of the forms and can capture aspects of the visual categorisation, they also miss some important features that differentiate the aesthetics of the individual forms. For example, Figure \ref{f:MENScores} shows the 25 lowest and highest ranked phenotypes for Euler morphology. The high values capture the ``worms'' category well, however the low scores mix low fitness scores and ``no grow\-th'' categories with the occasional highly ranked individuals in other categories. In summary, while these measures give reasonable indication of certain properties that relate to aesthetics, there are nuances in the dataset that simple image measures cannot capture: the motivation for why we turn to more advanced deep learning techniques later in this paper.

\subsection{Visualising Genotype Space}
\label{ss:genotypeVis}

\begin{figure*}
\hspace{0.53cm}\includegraphics[width=0.9\textwidth]{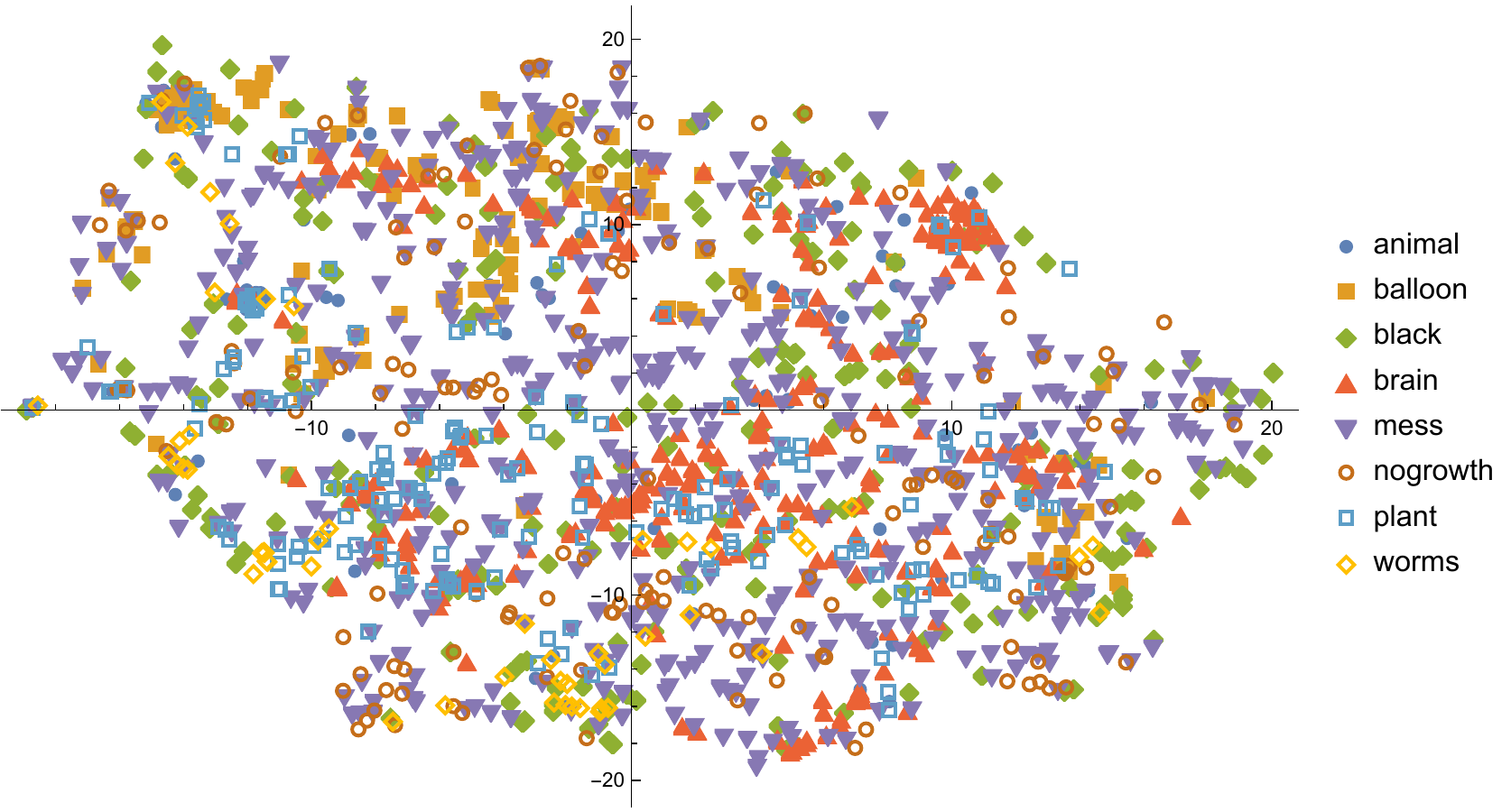}
\includegraphics[width=0.9\textwidth]{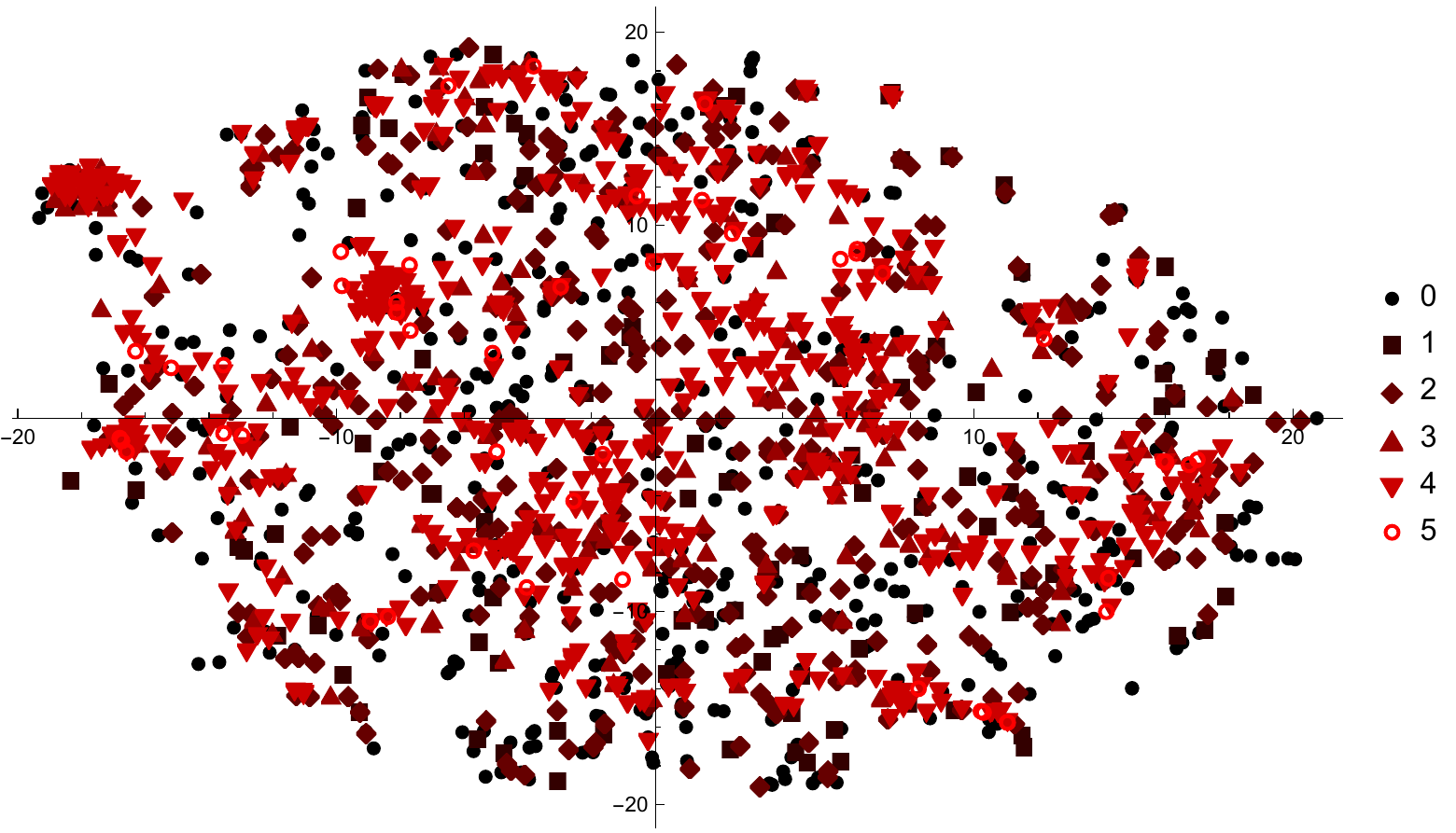}
\caption{Plot of genotype distribution in two dimensions using t-SNE. Individual genotypes are coloured by category (top) and by aesthetic rating score (bottom). Note that the original score range of 0-10 has been compressed into six bands for visual clarity. } \label{f:geneCat}
\end{figure*}

In order to better understand the design space, we tested dimension reduction algorithms on the dataset. Dimensionally reduced visualisations allow the designer to understand the high-dimensional search space more easily, for example to see clusters of specific form types or regions of high aesthetic value. Being able to predict the category or aesthetic value of a new genotype before the computationally expensive process of computing the phenotype from it can significantly reduce user fatigue and speed up the search process.

As shown in Figure \ref{f:geneCat}, the dimensionally reduced genotype space tends to have little visible structure. The figure shows each 12-dimensional genotype dimensionally reduced to two dimensions and colour and shape-coded according to category (top) and rating (bottom). In the case of rating, we reduced the eleven-point numeric scale to six bands for clarity. The figure shows the results obtained with the t-SNE dimension reduction with a perplexity of 60 and $\epsilon = 10$. Testing with other algorithms (PCA, UMAP and a Variational Autoencoder) did not result in significantly better visual results in terms of being able to visually distinguish categories or score clusters in the dimensionally reduced geneotype space visualisation.

Although some grouping can be seen in the figure, any obvious overall clustering is difficult to observe, particularly for the categories. While there is some overall structure in the score visualisation (some high ranked individuals are concentrated around the upper left quadrant), discerning any regions of high or low quality is difficult. In many cases, low and high ranked individuals map to close proximity in the 2D representation.

What this analysis reveals is that the genotype space is highly unstructured in relation to aesthetic concerns, making it challenging to easily evolve high quality phenotypes. The developmental nature of the generative system, which depends on physical simulation, means that small parameter changes at critical points can result in large differences in the resultant developed form.

\subsection{Visualising Phenotype Space}
\label{ss:phenotypeVis}
\begin{figure*}
\hspace{0.7cm}\includegraphics[width=0.9\textwidth]{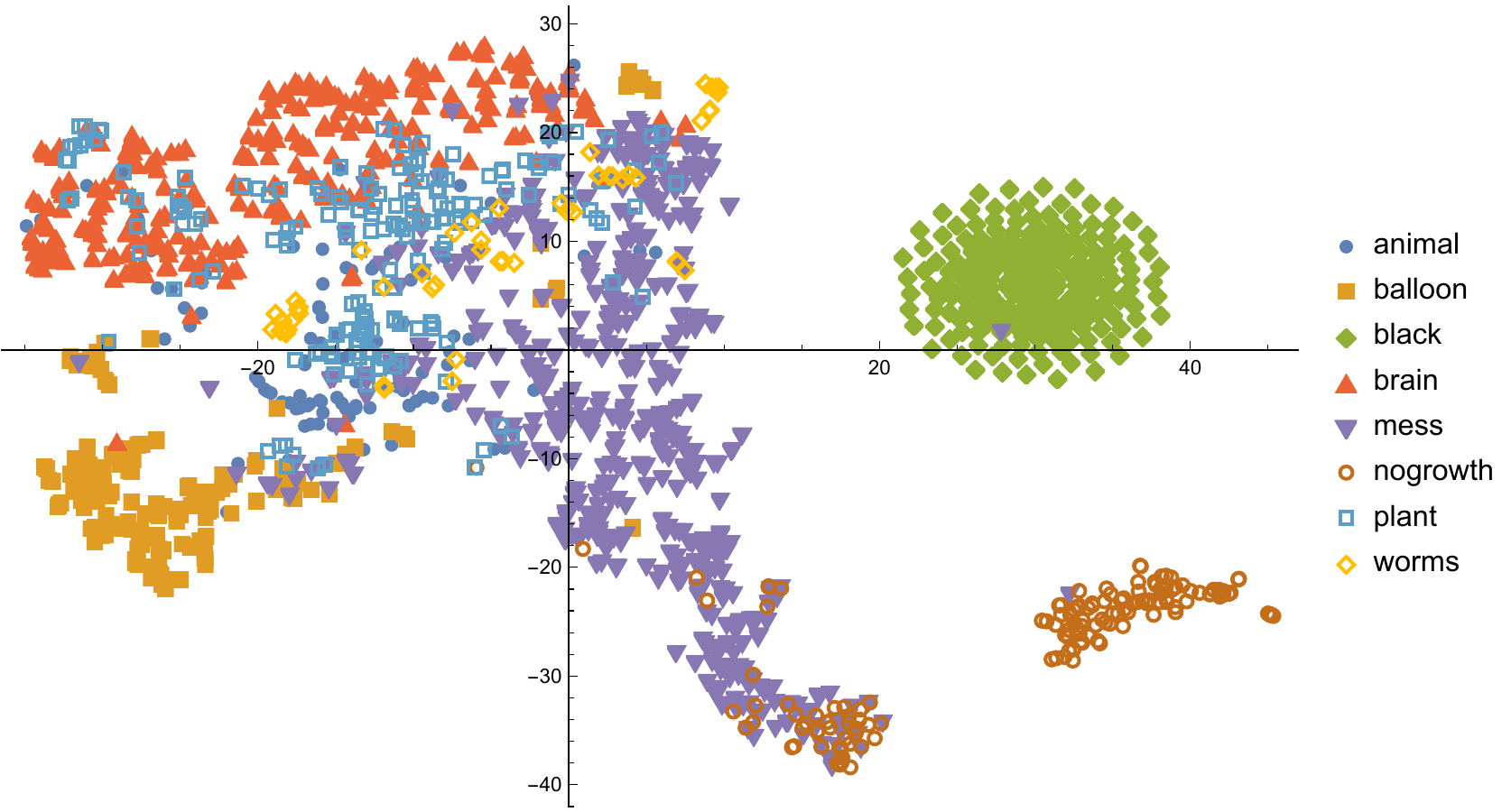}
\includegraphics[width=0.9\textwidth]{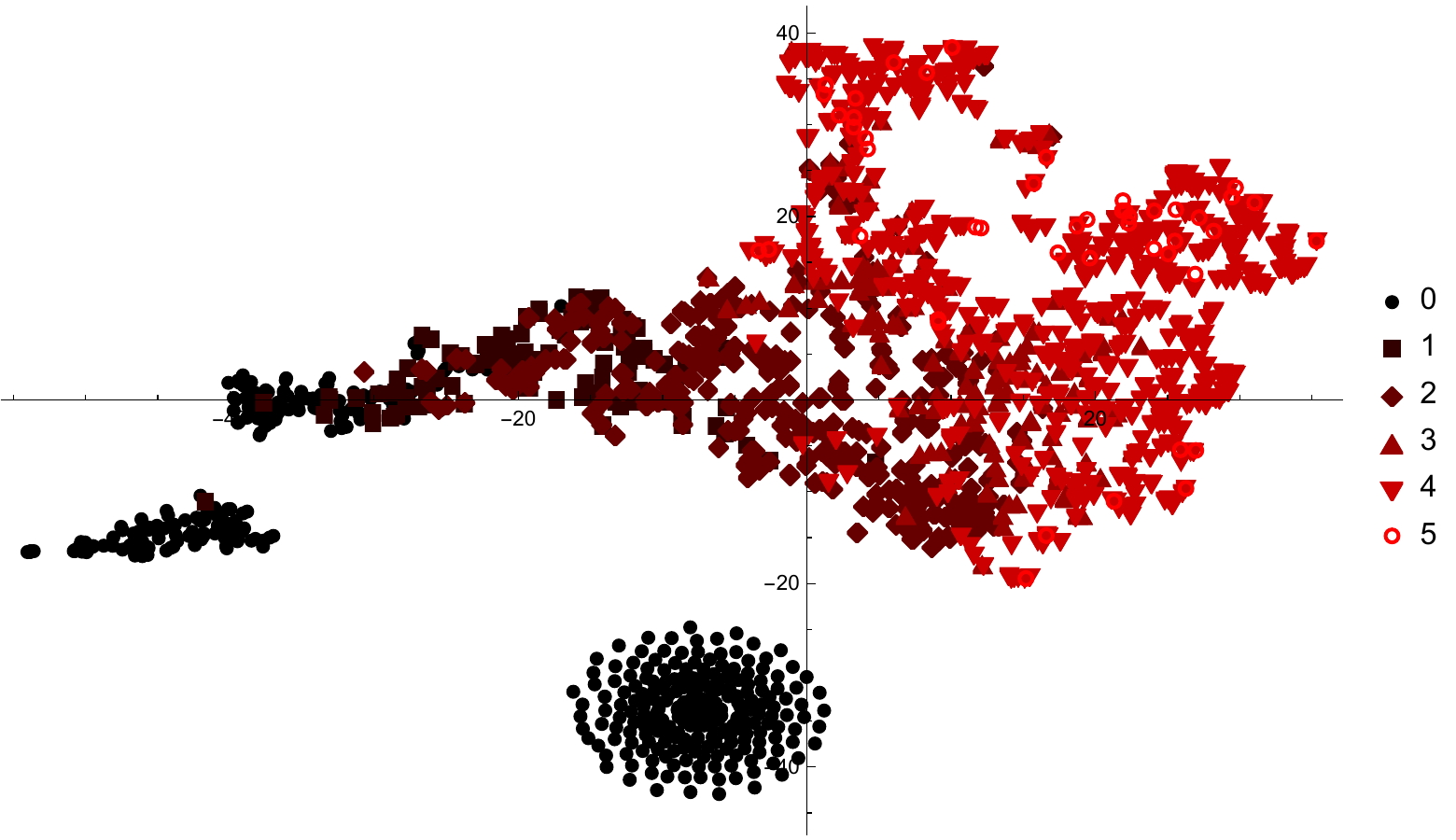}
\caption{Plot of phenotype distribution in two dimensions using t-SNE. Individual phenotypes are coloured by category (top) and by rating score (bottom). } \label{f:phenoCat}
\end{figure*}

To visualise the phenotype space we used the feature classification layer of the ResNet-50 convolutional neural network. Because ResNet was trained on 1.2 million images from the ImageNet dataset \cite{deng2009imagenet}, it is very good at identifying image features that humans also recognise.  Networks trained on the ImageNet classification tasks have been shown to work very well as off the shelf image extractors \cite{sharif2014cnn}, and show even better results when fine-tuned to datasets for the task at hand \cite{azizpour2015generic}. However, for this experiment we did not perform any additional tuning on the target dataset. The network produces a 2048-element vector based on the features of the input image. For these experiments we used an input image size of $512 \times 512$ pixels. The computed feature vector is then dimensionally reduced to create a two-dimensional visualisation of the feature space. Again, we used the t-SNE algorithm to reduce the dimensionality of the space with a perplexity of 20 and $\epsilon = 10$. A light linear pre-reduction was also performed before feeding the data into the t-SNE dimension reducer.

Figure \ref{f:phenoCat} shows the results for both the category (top) and score (bottom) classifications. As the figure shows, this time structure can be seen in the feature data. Classifications such as ``black'' and ``balloon'' are visible in specific regions. Similarly, the score distribution shows increasing values towards the upper-right quadrant in the visualisation. As with the genotype visualisation in Figure \ref{f:geneCat}, the 0-10 score range has been divided into 6 bands for visual clarity.

Such visualisations can assist artists in navigating and understanding the space of possibilities of their generative system, because they allow them to direct search in specific regions of phenotype (feature) space. A caveat here is that the dimension reduction process ideally needs to be reversible, i.e. that one can go from low dimensions back to higher if selection specific regions on a 2D plot.\footnote{In theory, a variational autoencoder could be trained to map genotype data to 2D images and if accurate enough, bypass the need for the original generative system altogether, although this would require a much larger training set to be aesthetically viable.}
As a minimum, it is possible to determine a cluster of nearby phenotypes in 2D space and seed the search with the genotypes that created them, employing methods such as hill climbing or nearest neighbour methods to search for phenotypes with similar visual features. 

We used the standard ResNet-50 convolutional neural network (CNN) to compute the feature vector for each image. So these visualisations suggest that CNNs can be used to recognise the features of abstract computational art images and differentiate between categories and aesthetic value of phenotypes as assigned by the artist, even without prior training on them. Hence they can be used, for example, as aesthetic fitness measures in an evolutionary system, or as filters to show fewer uninteresting images when using methods such as the IGA. The results shown in Figure \ref{f:phenoCat} led us to further explore the use of deep learning neural networks to assist in understanding individual aesthetic preferences, which we describe in Section \ref{s:learningPreferences}. 

\begin{figure}[tb]
\begin{center}
\includegraphics[width=0.45\textwidth]{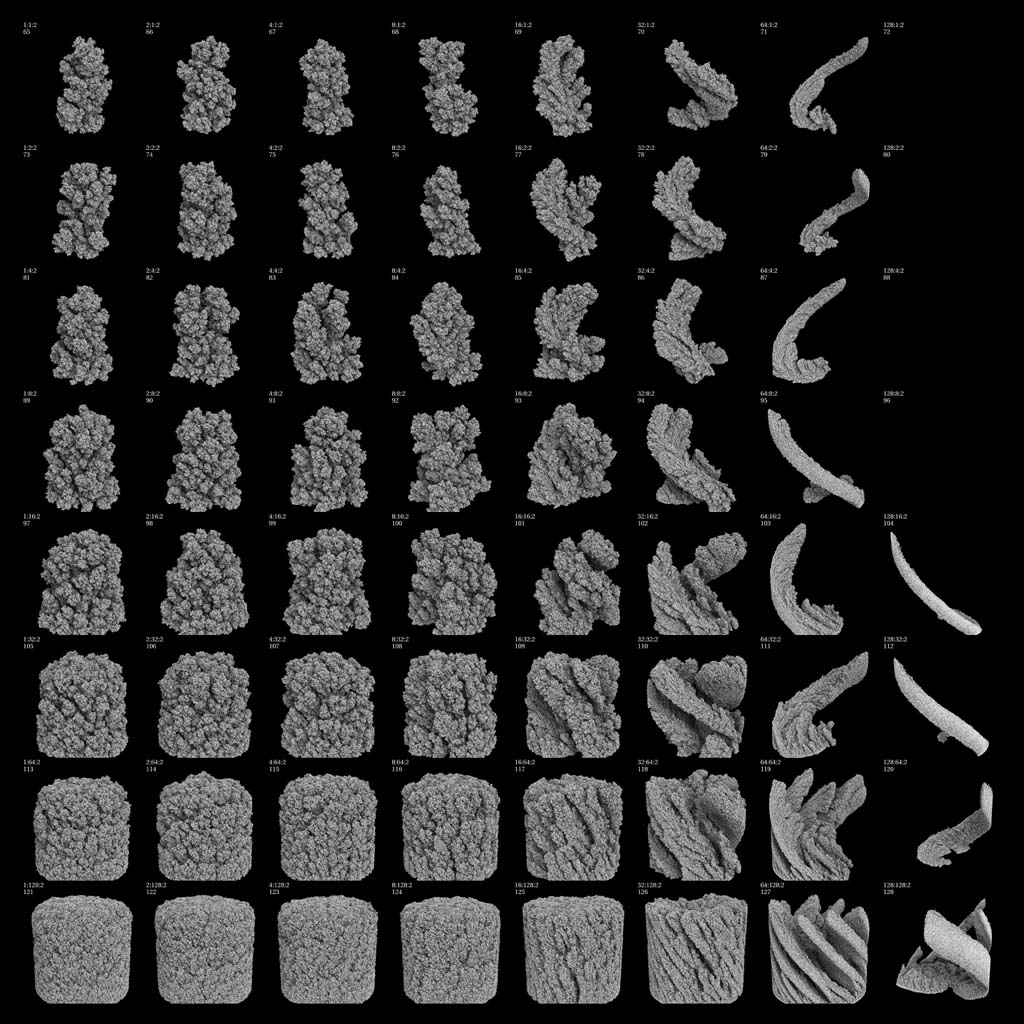}
\end{center}
\caption{Plot from Aggregation series showing effects of varying genotype parameters} \label{f:aggregationPlot}
\end{figure}

\subsection{Parameter Searching and Interpolation}
\label{ss:parameterSearch}

In early work, such as Lomas' Aggregation \cite{Lomas2005} and Flow \cite{Lomas2007} series, the artist would create plots showing how the phenotype changes depending on parameter values of the genotype. An example of such a plot is shown in Figure \ref{f:aggregationPlot}. In these systems the genotype has a very low number of dimensions, typically just two or three parameters, which allowed a dense sampling of the space of possibilities by simply independently varying each parameter in the genotype over a specified range, running the generative system with each set of parameter values, and plotting the results in a chart with positions for each image based on the genotype parameters. One intuition from these plots is that the most interesting, rich and complex behaviour often happens at transition points in the genotype space, where one type of characteristic behaviour or ``category'' changes into another. This can be seen in Figure \ref{f:aggregationPlot} where the forms in the 6th and 7th columns are particularly richly structured. These changes occur at parameter settings where the generative system was at a transition state between stability (to the left) and instability (to the right).

However, as the number of dimensions increases performing a dense sampling of the genotype space runs into the ``Curse of Dimensionality'' \cite{Bellman1961,Donoho2000}, where the number of samples needed increases exponentially with the number of parameters. Even if enough samples can be taken, how to visualise and understand the space becomes difficult and concepts such as finding the nearest neighbours to any point in the parameter space become increasingly meaningless \cite{Marimont1979}. One potential approach to make sense of higher dimensional spaces is to categorise different phenotypes created by the system. By defining categories for phenotypes we can express searching for transition points in a meaningful way as being the places in genotype space where small changes in the genotype result in changing from one phenotype category to another.

\section{Learning an Artist's Aesthetic Preferences}
\label{s:learningPreferences}

A ResNet-50 CNN was re-trained and tested with with the same dataset of 1,774 images with ratings and categories as described above, using 1,421 images in the training dataset and 353 images in the validation dataset. To use the network as a predictor, we removed the last three layers of the standard network and added new classification and softmax layers. We generated two separate networks, one for artist-defined categories and one for the aesthetic rating scores. The network output is a set of probabilities for each category (for the category network) or score (for the score network).

For the category network, training was performed using fast.ai’s \cite{howard2018fastai} \texttt{fit\_one\_cycle} method for 4 epochs with a learning rate of 0.001 followed by an additional 4 epochs with a learning rate of 0.00001. The score network was also trained using using the \texttt{fit\_one\_cycle} method for 4 epochs with a learning rate of 0.01 followed by an additional 4 epochs with a learning rate of 0.001.

Re-training the final classifier layers of ResNet-50 created a network that matched Lomas' original categories in the validation set with an accuracy of 87.0\%. We also looked at the confidence levels for the predictions, based on the difference between the network's probability value for the predicted category and the probability level of the highest alternative category.

\begin{table}
\caption{ResNet-50 accuracy levels for different confidence quartiles.}
\label{tab:accuracyConfidence}
\begin{tabular}{ll}
\hline\noalign{\smallskip}
Confidence quartile & Prediction accuracy\\
\noalign{\smallskip}\hline\noalign{\smallskip}
75\% to 100\% & 97.1\%\\
50\% to 75\% & 97.9\%\\
25\% to 50\% & 90.5\%\\
0\% to 25\% & 67.6\%\\
\noalign{\smallskip}\hline
\end{tabular}
\end{table}

Table \ref{tab:accuracyConfidence} shows how the prediction accuracy varies depending on the confidence levels. The network has a reliability of over 97\% for the images in the top two confidence quartiles, with 69\% of the incorrect categorisations being in the lowest confidence quartile. A visual inspection of images in the lowest confidence quartile confirmed that these were typically also less clear which category an image should be put in to a human observer.

\begin{figure}
\begin{center}
\includegraphics[width=0.45\textwidth]{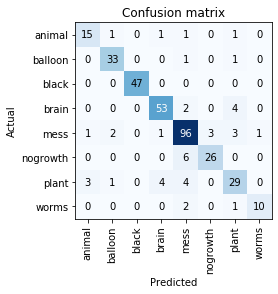}
\end{center}
\caption{Confusion matrix for ResNet-50 (phenotype space) category categorisor} \label{f:resnetConfusionMatrix}
\end{figure}

The confusion matrix in Figure \ref{f:resnetConfusionMatrix} shows that the predictions appear to be consistently good across all categories, with the majority of items in each category predicted correctly. The most confused categories were ``mess'' and ``no\-growth'', both of which indicate forms that are considered by the artist to be aesthetic failure cases and sometimes look quite similar. In particular, the ``nogrowth'' category is quite distinctive and was relatively easy to classify using the image analysis methods discussed in Section \ref{ss:imageAnalysis} (see some examples in Figure~\ref{f:MENScores}).

As explained, a separate ResNet-50 network was trained against the aesthetic ranking using values from 0 to 10 that Lomas had given the forms. This resulted in a network that predicted the ranking of images in the validation set with a root mean square error of 0.716. Given that these ranking are subjective evaluations of images that often have very similar appearance this appears to be a high level of predictive accuracy.

\subsection{Genotype Space}
\label{ss:genotypeSpace}

Given the promising results in phenotype space, we next turned to see if neural networks could be used to predict the artist assigned aesthetic scores and categories directly from the genotype. Recall that the dimension reduction visualisation discussed in Section \ref{ss:genotypeVis} had not shown any obvious clustering in two dimensions. The dataset was tested to see whether predictions of the phenotype category and aesthetic rank could be obtained from genotype parameters. This is desirable as good predictions of phenotype from the genotype values could directly aid exploration of the space of possibilities.
Additionally, as the generation process from genotype to phenotype is computationally expensive, knowing promising areas of genotype space to explore can potentially save a lot of time where the artists waits for phenotpyes to be generated.
Techniques such as Monte Carlo methods can be used to choose new candidate points in genotype space with specified fitness criteria. We could use the predictions to generate plots of expected behaviour as genotype parameters are varied that could help visualise the phenotype landscape and indicate places in genotype space where transitions between phenotype classes may occur. If meaningful gradients can be calculated from predictions, gradient descent could be used to directly navigate towards places in genotype space were one category is predicted to change into another and transitional forms between categories may exist.

Fast.ai \cite{howard2018fastai}, a Python machine learning library to create deep learning neural networks, was used to create neural net predictors for the category and aesthetic rank using genotype values as the input variables. The fast.ai Tabular model was used, with a configuration of two fully connected hidden layers of size 200 and 100. The same training and validation sets were used as previously.

Using these neural nets we achieved an accuracy of 68.3\% for predictions of the category, and predictions of the aesthetic rank had a root mean square error of 1.88. These are lower quality predictions that we obtained with the ResNet-50 classifier using the phenotype, but this is to be expected given that the ranking and categorisation are intended to be evaluations of the phenotype and were done by Lomas looking at the images of the phenotype forms. The results are also confirmed by the dimensionally-reduced visualisations presented in Section \ref{ss:designSpace}.

\begin{figure}
\begin{center}
\includegraphics[width=0.45\textwidth]{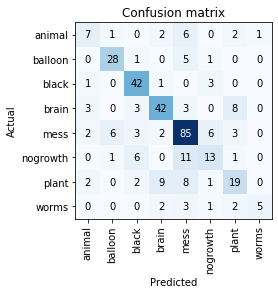}
\end{center}
\caption{Confusion matrix for Tabular (genotype space) categoriser} \label{f:tabularConfusionMatrix}
\end{figure}

The confusion matrix for the category predictions is sh\-own in Figure \ref{f:tabularConfusionMatrix}. Similarly to the results with the ResNet-50 (phenotype space) categoriser, the ``mess'' and ``nogrowth'' categories are often confused, but with the genotype space categoriser the ``plant'' and ``brain'' categories are also quite frequently confused with each other. This suggests that it might be worth generating more training data for the ``brain'' and ``plant'' categories to improve predictive accuracy, but could also be an indication that the ``plant'' and ``brain'' categories are closely connected in genotype space. As the visualisation in Figure \ref{f:phenoCat} shows, there is significant overlap between these two categories in the dimensionally reduced feature space.

The previous version of Lomas' \emph{Species Explorer} software used a simple k-Nearest Neighbours (k-NN) method to give predictions of phenotype based on genotype data \cite{Lomas2014}. Testing with the same validation set, the k-NN method predicts categories with an accuracy of 49.8\%, and the aesthetic rank with a mean square error of 2.78. The genotype space neural net predictors give significantly better predictions than the k-NN predictor.

\begin{figure*}
\includegraphics[width=\textwidth]{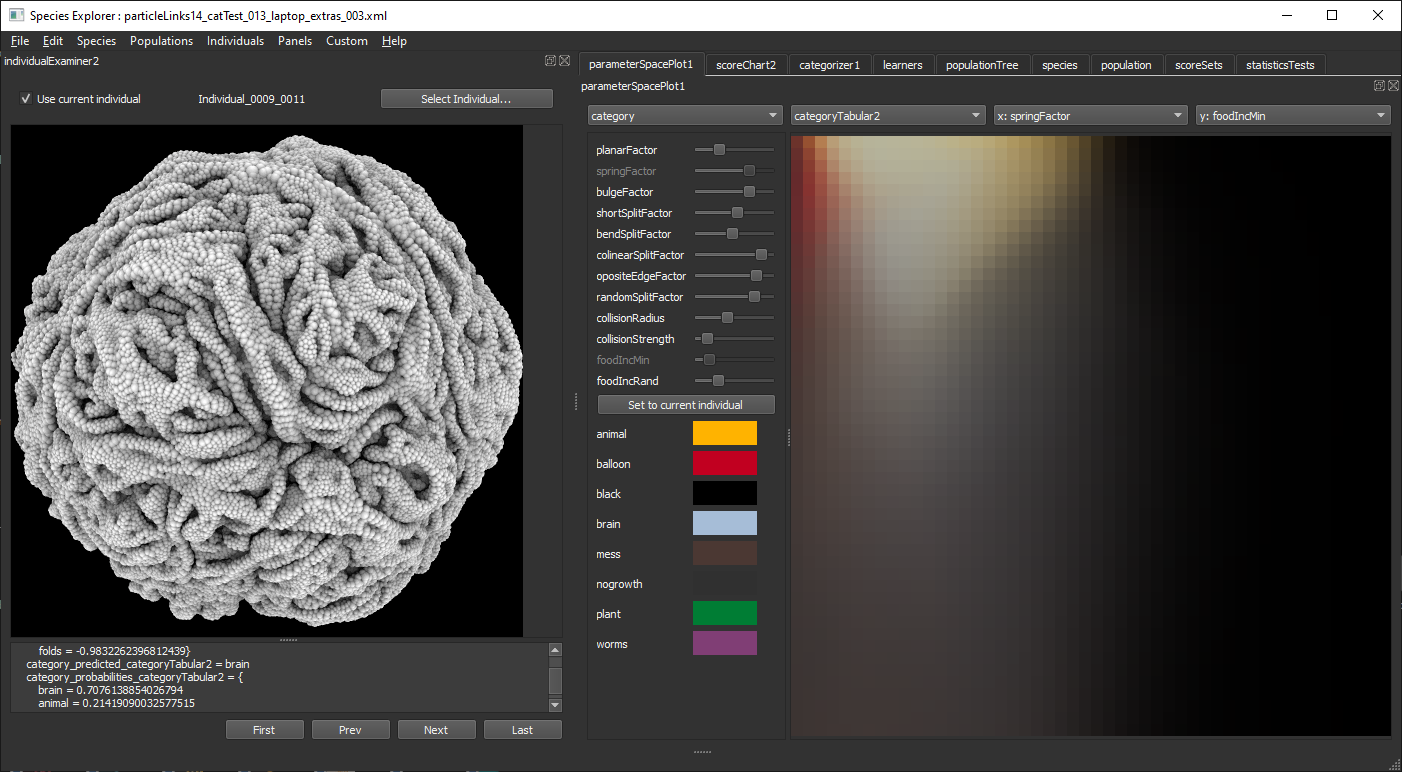}
\caption{Species Explorer user interface, showing 2D cross-section plots through genotype space using a neural net to predict the phenotype category at new positions in genotype space} \label{f:speciesExplorerGenotypeSpacePlot}
\end{figure*}

A new feature was added into Species Explorer that uses the genotype neural network predictors to generate 2D cross-section plots through genotype space, showing the predicted categories or rank at different coordinate positions, see Figure \ref{f:speciesExplorerGenotypeSpacePlot}. As can be seen, these plots predict a number of potential places where transitions between categories may occur, which could lead the artist to intuitively explore new regions of the genotype space.

\section{Discussion}
\label{ss:discussion}

The results of incorporating this new search feature into an artist's creative workflow indicate that deep learning based neural networks appear to be able to achieve good levels of accuracy when predicting the phenotype categories and aesthetic rank evaluations made by Lomas in the test dataset. The best predictions were achieved with ResNet-50, a pre-trained convolutional neural network designed for image re\-cog\-nition, using phenotype image data as the input. Additionally, we achieved potentially useful levels of prediction from genotype data using the fast.ai library's Tabular model to create a deep learning neural net with two fully connected hidden layers. Predictions based on the genotype rather than the phenotype are particularly interesting as they allow navigation directly in genotype space to suggest new points to sample.

The main motivation for using IGAs is that the fitness function is unknown, or may not even be well defined because the artist's judgement changes over time. The use of the neural networks in this work can be seen as trying to discover whether there is a function that matches the artist's aesthetic evaluations with a useful level of predictive utility. If such a function can be found it could be used in a number of ways, such to use monte carlo sampling along with providing a fitness function for conventional evolutionary algorithms. If the discovered fitness function is sufficiently simple (such as being unimodal) methods like hill climbing may be appropriate.

Lomas has been using a k-Nearest Network in Species Explorer to give prediction based on position in genotype space. As the numbers of dimensions increase k-NN performance generally becomes significantly less effective \cite{Marimont1979}, while deep neural networks can still be effective predictors with higher dimensional inputs. This means that deep neural networks have the potential to allow useful levels of prediction from genotype space in systems with high numbers of genotype parameters.

It is likely that with more training data we will be able to improve the predictions from genotype space. This raises the possibility for a hybrid system: if we have a convolutional neural network that can achieve high levels of accuracy from phenotype data we could use this to automate creation of new training data for a genotype space predictor. In this way, improving the ability of a genotype space based predictor may be at least partially automated.

There is a lot of scope for trying out different configurations of deep neural networks for genotype space predictions. The choice of a network in this study, with two fully connected hidden layers with 200 and 100 neurons, was simply based on the default values suggested in the fast.ai documentation of their Tabular model. A hyper-parameter search would likely reveal even better results.

An important part of this process is how to make ranking and categorisation as easy for a creative practitioner as possible. The aim should be to allow the artist to suggest new categories and ways of ranking with as few training examples as are necessary to get good levels of prediction. It should also facilitate experimentation, making the process of trying out new ways of ranking and different ways of categorising behaviour as simple as possible.

In both authors' experience, it is often only after working with a generative system for some time, typically creating hundreds of samples, that categories of phenotype behaviour start to become apparent. This means that manually categorising all the samples that have already been generated can become significantly laborious. This has meant that although Lomas' Species Explorer software allows phenotype samples to be put into arbitrary categories, and data from categorisation can be used to change fitness functions used to generate new samples, for the majority of systems Lomas has created he hasn't divided phenotype results into categories and has relied on aesthetic rank scores instead. This is one area where pre-trained network classifiers, such as ResNet-50, may be useful. If we can reliably train a neural network to classify different phenotypes with only a small amount of training data it could make the process of creating and testing different ways of categorising phenotypes significantly easier.

\begin{figure*}
\includegraphics[width=\textwidth]{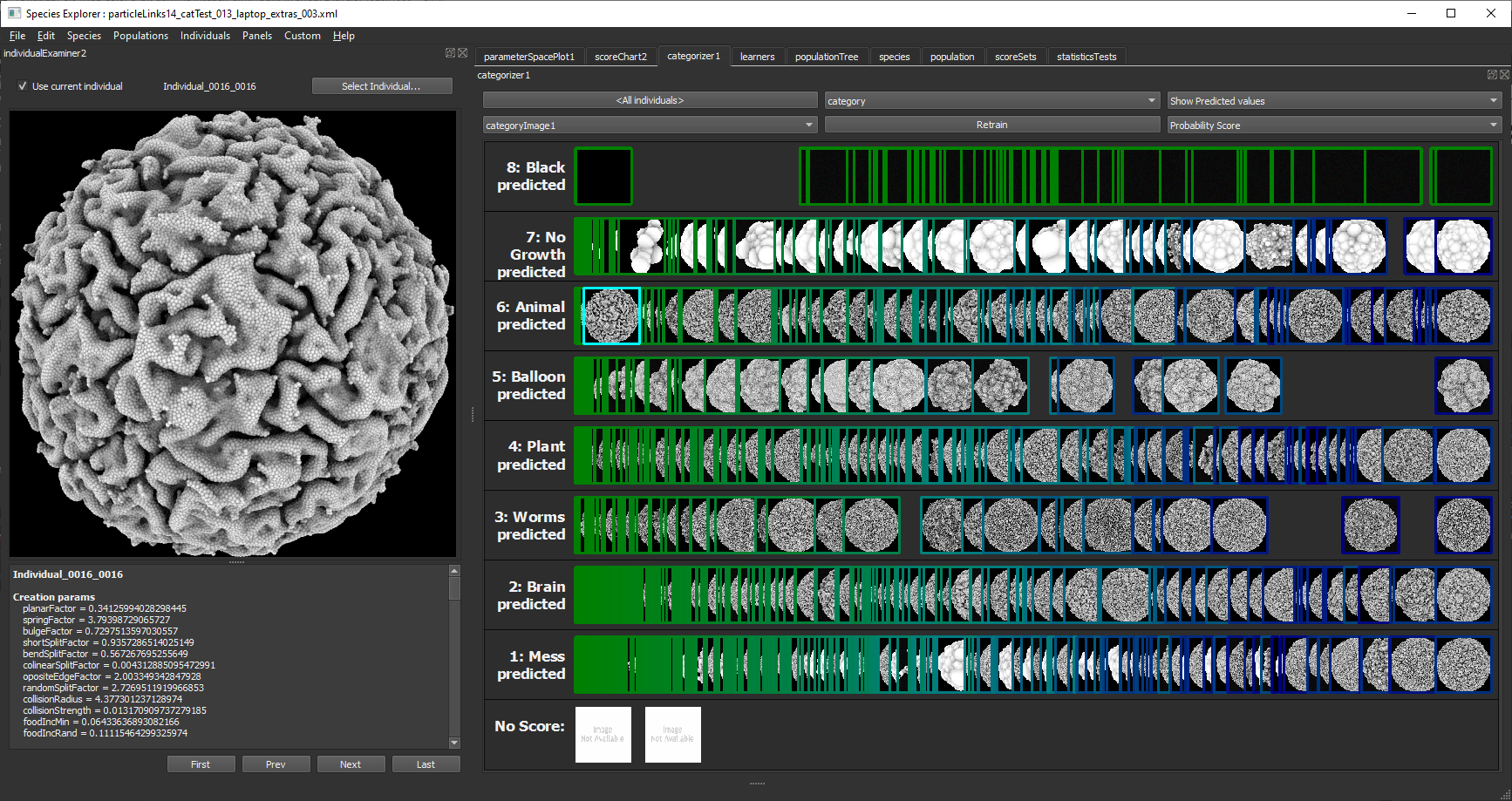}
\caption{Species Explorer user interface, showing predicted categorisations from a ResNet-50 network. The items are ordered based on the confidence levels for the predicted category, with the highest confidence level predictions to the left of each group} \label{f:speciesExplorerProbabilityScores}
\end{figure*}

To test this, we modified the existing user interface in Species Explorer so that predictions of how a classifier would divide data into classes can be shown, together with placement and colouring of the outlines of thumbnails based on the confidence levels of predictions, see Figure \ref{f:speciesExplorerProbabilityScores}. This allows a simple evaluation of the quality of prediction, and helps indicate samples that might be good to add to the training set (such as incorrect predictions that the classifier has done with high confidence) to improve the quality of predictions.

The tests with dimensionally reduced plots in phenotype space using t-SNE on the feature vectors of ResNet-50 appear to show meaningful structure which may be useful to help divide samples into categories. In particular, this technique may be useful both to help initial categorisation, broadly dividing samples in phenotype space into categories, and to help sub-divide existing categories that the user wants to explore separating into different classes. The use of plots such as these may actively help experimentation, allowing the creative users to modify existing classification schemes and quickly try out different ideas of how to categorise phenotypes.

\section{Conclusions and Future Work}
\label{s:conclusion}

The aim of this research was to progress machine-assisted aesthetic judgement based on an artist's personal aesthetic preferences. We worked with an established artist's work to give ecological validity to our system and its results. While the results are specific to an individual artist, it is worth emphasising that the methods discussed generalise to any multi-parameter generative system whose phenotypes can be expressed as 2D images. Indeed, the Species Explorer software separates the creative evolution process from the actual generative system, allowing Species Explorer to work with \emph{any} parameter based generative system. 

The research presented here shows that deep learning neural networks can be useful to predict aesthetically driven evaluations and assist artists to find phenotypes of personally high aesthetic value. As discussed, these predictors are useful to help explore the outputs of generative systems directly in genotype space. Incorporating these methods into IGA systems can reduce the amount of time wasted looking at aesthetically poor or previously seen phenotypes, helping to eliminate  user fatigue in the conventional IGA. 

There is still more research to be done however.
More testing is now needed to see how productive this is in practice when working with systems that often have high dimensional parameter spaces. We have shown the neural networks can categorise and rank phenotypes with a high accuracy in a specific instance, our next stage of research  is to see if this approach generalises to other artists and their personal aesthetics. A challenge is getting access to high quality data. We have made our dataset publicly available and encourage other artists to allow public access to their own work in order to advance research in this field.

\section*{Conflict of interest}
The authors declare that they have no conflict of interest.

\begin{acknowledgements}
We would like to thank the reviewers for their helpful comments in improving the final version of this paper.
This research was supported by an Australian Research Council grant FT1701\-00033 and a Monash University International Research Visitors Collaborative Seed Fund grant.
\end{acknowledgements}

\bibliographystyle{spmpsci}      

\bibliography{mainRefs}

\end{document}